\documentclass{article}

\usepackage[final]{neurips_data_2021}

\usepackage[utf8]{inputenc} 
\usepackage[T1]{fontenc}    
\usepackage{hyperref}       
\usepackage{url}            
\usepackage{booktabs}       
\usepackage{amsfonts}       
\usepackage{nicefrac}       
\usepackage{microtype}      
\usepackage{graphicx}

\title{AI and the Everything in the Whole Wide World Benchmark}

\author{
Inioluwa Deborah Raji\\
Mozilla Foundation, UC Berkeley\\
rajiinio@berkeley.edu 

\And

Emily M. Bender\\
Department of Linguistics\\
University of Washington

\And

Amandalynne Paullada\\
Department of Linguistics\\
University of Washington

\And

Emily Denton\\
Google Research

\And

Alex Hanna\\
Google Research

}

\usepackage[suppress]{color-edits}
\addauthor{EMB}{orange}
\addauthor{AP}{purple}
\addauthor{AH}{teal}
\addauthor{ED}{blue}
\addauthor{DR}{violet}

\usepackage{graphics}
\usepackage{wrapfig}

\hypersetup{
    colorlinks=true,
    linkcolor=blue,
    filecolor=magenta,      
    urlcolor=blue,
    citecolor=black
}

\begin{document}

\maketitle

\begin{abstract}

There is a tendency across different subfields in AI to valorize a small collection of influential benchmarks. These benchmarks operate as stand-ins for a range of anointed common problems that are frequently framed as foundational milestones on the path towards flexible and generalizable AI systems. State-of-the-art performance on these benchmarks is widely understood as indicative of progress towards these long-term goals. In this position paper, we explore the limits of such benchmarks in order to reveal the construct validity issues in their framing as the functionally ``general'' broad measures of progress they are set up to be.

\end{abstract}

\section{Introduction}

In the 1974 Sesame Street children's storybook \emph{Grover and the Everything in the Whole Wide World Museum} \citep{Grover:74}, the Muppet monster Grover visits a museum claiming to showcase ``everything in the whole wide world''. Example objects representing certain categories fill each room. Several categories are arbitrary and subjective, including showrooms for ``Things You Find On a Wall'' and ``The Things that Can Tickle You Room''. Some are oddly specific, such as ``The Carrot Room'', while others unhelpfully vague like ``The Tall Hall''. When he thinks that he has seen all that is there, Grover comes to a door that is labeled ``Everything Else''. He opens the door, only to find himself in the outside world.

As a children's story, Grover's described situation is meant to be absurd. However, in this paper, we discuss how a similar faulty logic is inherent to recent trends in artificial intelligence (AI)\,---\,and specifically machine learning (ML)\,---\,evaluation, where many popular benchmarks rely on the same false assumptions inherent to the ridiculous ``Everything in the Whole Wide World Museum'' that Grover visits. In particular, we argue that benchmarks presented as measurements of progress towards general ability within vague tasks such as ``visual understanding'' or ``language understanding'' are as ineffective as the finite museum is at representing ``everything in the whole wide world,'' and for similar reasons\,---\,being inherently specific, finite and contextual.

Benchmarks like GLUE \citep{wang2018glue} or ImageNet \citep{imagenet_cvpr09} are often elevated to become definitions of the essential common tasks to validate the performance of any given model. As a result, often the claims that are justified through these benchmark datasets extend far beyond the tasks they are initially designed for, and reach beyond even the initial ambitions for development. Despite a presentation and acceptance as markers of progress towards general-purpose capabilities, there are clear limitations of these benchmarks. In fact, the reality of their development, use and adoption indicates a \emph{construct validity} issue, where the involved benchmarks\,---\,due to their instantiation in particular data, metrics and practice\,---\,cannot possibly capture anything representative of the claims to general applicability being made about them. In this paper, we  illuminate the over-reliance on the historical Common Task Framework (CTF) for machine learning as it inappropriately evolved into what we understand today to be these benchmarks claiming to assess general capabilities. We do not deny the utility of such benchmarks, but rather hope to point to the risks inherent in their framing. Our goal is to dig into the details of what it means for us as a community to continue propping up certain benchmarks with the rhetoric of ``general'' performance.  

\section{Background}

\subsection{Striving for Generality} 

The notion of a ``general-purpose'' AI can be traced, in part, back to the early days of the field. In the late 1950s, \cite{newell1959report} proposed the ``General Problem Solver'' in an attempt to create a system that could solve a range of problems, as well as develop a general theory of problem solving. While this project ultimately failed, it helped establish general intelligence as a goal within the field of AI. As \cite{Adam98} notes, the failure of the project was at the time attributed primarily to implementation issues, rather than the perception of any flaw in the formulation of a disembodied and general-purpose problem solver as a meaningful goal.

Researchers now often use the term ``generality'' to refer to the development of AI systems that are flexible in nature, demonstrating competence on a wide range of tasks and in a wide range of settings, hoping to mirror the adaptive cognitive abilities humans are perceived to possess \citep{Shevlin2019}. While discussions of ``generality'' might evoke the goal of artificial general intelligence \citep{Voss2007,Pennachin2007}, we set that aside in this paper and focus instead on the more immediate (and logically prior) issue of claims of generality for a specific type of cognitive skill or capability (e.g., vision or language), supposedly independent of context or application domains.

Within specific sub-fields and more pragmatic settings, researchers will often strive for the development of general-purpose systems that capture a breadth of functionality or knowledge. The development of general-purpose feature embeddings has also become a central research focus within machine learning communities. Here, the goal is the development of representations\,---\,through either unsupervised or supervised means\,---\,that can generalize (with minimal fine-tuning) to a wide range of other tasks they were not specifically developed for  \cite[e.g.][]{huh2016what, du2020, bommasani2021opportunities}.

\subsection{A Brief History of Benchmarking Practice in AI}

Some assess AI progress by asking scientific and philosophical questions on the notion of intelligence, thinking of AI as a method for modeling some general cognitive function. Others, particularly those in the machine learning field, evaluate performance based on the success of a model’s utility within some practical applications. Although the former goals sparked the aspiration to measure general cognitive capabilities, the latter application-based objectives are what data benchmarking has been historically designed to communicate.

In this paper we describe a \emph{benchmark} as a particular combination of a dataset or sets of datasets (at least test data, sometimes also training data), and a metric, conceptualized as representing one or more specific tasks or sets of abilities, picked up by a community of researchers as a shared framework for the comparison of methods. The \emph{task} is a particular specification of a problem, as represented in the dataset.\footnote{This corresponds to \citeauthor{Schlangen2020TargetingTB}'s \citeyearpar{Schlangen2020TargetingTB} notion of extensional definition of tasks; see \S\ref{sec:limit-design}.} A \emph{metric} is a way to summarize system performance over some set or sets of tasks as a single number or score. The metric provides a means of counting success and failure at the level of individual system outputs and summarizing those counts over the full dataset. Models obtaining the most favourable scores on the metrics for a benchmark are considered to be ``state-of-the-art'' (SOTA) in terms of performance on the specified task. Here, we present a stylized history that's far from comprehensive, but attempts to provide the shape of the development of predecessors of current machine learning benchmarking practice.

Before benchmarks were employed for algorithm selection in machine learning, they were used for ``computer selection''\,---\,i.e.\ the task of running a benchmark program on multiple computers to determine which machine was most suitable for purchase. In 1962, Auberbach Corporation's Standard EDP Reports were developed as a way to directly assess the performance of machines on specific prototype tasks, to avoid the tedious process of individually weighing the quality of the various features of computers offered by different vendors. By 1976, benchmarking was a fairly common practice in computing, with standard libraries of mock software programs created for the purpose of testing specific subroutines for core functional tasks (i.e.\ file management, word processing, etc.) being developed and used by the U.S. government as part of the assessment involved in their procurement process for computer systems \citep{lewis1985evolution}.

Several linguists \citep[e.g.][]{liberman2010fred, churchEmergingTrendsTribute2018} have traced the influence of Fred Jelinek at IBM and Charles Wayne at DARPA as crucial protagonists in the development of a common framework for the quantitative assessment of computational linguistic tasks, notably within speech recognition and machine translation. Beginning in the mid-1980s, Jelinek had been a driver in developing what he had called the Common Task Framework (CTF). The Common Task Framework, the precursor to modern data benchmarking in AI, was initially introduced in the 1980s as a response to Wendell Pierce's influential assertion of a field ``being deceived by the glamour of (a would-be) theory, rather than actual performance''  \citep[p.17]{donoho50YearsData2017}. The primary elements of the CTF include (a) A publicly available training dataset with a list of feature measurements and a class label for that observation; (b) A set of enrolled competitors whose common task is to infer a class prediction rule from the training data; and (c) A scoring referee, to which competitors can submit their prediction rule \citep[p.\ 572]{donoho50YearsData2017}. With DARPA's sponsorship and IBM's data, the CTF soon became widely adopted as a common mode of research in machine translation and speech recognition. Rather than experimental results that were updated ``twice a year'', researchers could evaluate their experiments every hour \citep{churchEmergingTrendsTribute2018}, and, although major breakthroughs were still rare, claims to incremental improvement enabled by this ``quantitative comparison of alternative algorithms on a fixed task'', ensured that researchers could track steady progress regardless \citep{liberman2010fred}. Soon, similar NLP-related initiatives followed\,---\,such as Shared Task Evaluation Campaigns (STECs) \citep{belz2006shared}, Message Understanding Conferences (MUCs) \citep{grishman1996}, and Text REtrieval Conferences (TRECs) for information retrieval.\footnote{\url{https://trec.nist.gov}} Later, STECs within NLP coalesced around machine translation (NIST-MT)\citep{przybocki2009nist} and word sense disambiguation (SENSEVAL)\citep{kilgarri1998senseval}.

Parallel developments also occurred in other fields. The Facial Recognition Technology (FERET) program for automated facial recognition, inaugurated in 1993 \citep{phillips2000feret}, was one of the first to bring a CTF type evaluation approach to the computer vision. The National Institute of Standards (NIST) reports that, prior to the publication of their FERET database,  ``[o]nly a few of these algorithms reported results on images utilizing a common database let alone met the desirable goal of being evaluated on a standard testing protocol that included separate training and testing sets. As a consequence, there was no method to make informed comparisons among various algorithms'' \citep[p. 2]{phillips2000feret}. 
Meanwhile, modern reinforcement learning algorithms have traditionally anchored to a culture of benchmarking, often evaluated on gaming environments like Atari, Starcraft, Dota2, and Go. One of the first games to widely capture the imagination of AI researchers was chess which became a fixture of US-Soviet Cold War competition and a stand-in for national technological achievement. \citep[p.\ 6]{Ensmenger:12}. 
More generally, the UC Irvine Machine Learning Repository (UCI) \citep{Dua:2019} was created in 1987 as a response to many calls for machine learning to have a centralized location for data in machine learning \citep{radinDigitalNativesHow2017} and has morphed into a repository for a wide array of tasks and data.

\subsection{Construct Validity}

If reproducibility and reliability are about the precision and thus the reliable repeatability of a finding, then challenges with validity for machine learning would be focused on how the field approaches the question of accuracy\,---\,namely, how closely our evaluations hit the mark in appropriately characterizing the actual anticipated behaviour of the system in the real world or progress on stated motivations and goals for the field. The accuracy of these evaluations is thus not just a question of reporting consistency but an actual design task involving reflection on what the objectives of an effective model involve, in addition to conscious decision making on how to best represent these desired outcomes in evaluation metrics, data and methodology. 
In particular, \emph{construct validity} is an external validity issue related to how well an experimental setting relates to a research claim, or, in the context of machine learning, how well the benchmark dataset, and associated metrics of evaluation, represents a task \citep{o1998empirical}. It concerns how well designed\,---\,or rather, how well \emph{constructed}\,---\,the experimental setting is in relation to the research claim. Validity issues have long been acknowledged broadly by the machine learning community as a particularly difficult task in benchmark design and development \citep{mitchell2021ai, Jac:Wal:21, malik2020hierarchy}\,---\,however as \citeauthor{bowman2021will} note, it remains a generally neglected issue as ``[t]his criterion is difficult to fully formalize, and we know of no simple test that would allow one to determine if a benchmark presents a valid measure of model ability'' \citep[p.\ 3]{bowman2021will}.

\section{Attempting to Benchmark General Capabilities}
\label{sec:general_benchmark}

We now examine a particular style of benchmark dataset\,---\,and the accompanying practices of use\,---\,that has gained in popularity in recent years. These benchmarks have been claimed to embed notions of generality, both in their presentation by the creators and the manner in which they are adopted and used by the machine learning community. Even when the creators of such benchmarks do not explicitly purport to be establishing benchmarks for ``general intelligence'', community practices and overall hype have dramatized what it means for a model to perform well on these benchmarks. The inspiration for the setup of these benchmarks is often explicitly linked to the general-purpose capabilities  and breadth of knowledge humans possess. In doing so, the ethos that surrounds these datasets often extends beyond a reasonable scope of interpretation and influence, especially given the limitations inherent in their construction.
 
In this paper, we argue that the aim of measuring general-purpose capabilities (i.e.\ goals such as general-purpose object recognition, general language understanding or domain-independent reasoning) cannot be adequately embodied in data-defined benchmarks. We observe that current trends inappropriately extend the Common Task Framework (CTF) paradigm to apply to the ``task'' of performance in the abstract, distinct from real world objectives or context. Historically, the CTF was developed precisely to introduce practically-oriented and tightly-scoped AI tasks\,---\,namely, automatic speech recognition (ASR) or machine translation (MT)\,---\,where the required validation is whether the benchmark accurately reflects the practical task being asked of the computer in its real-world context \citep{donoho50YearsData2017}. This new wave of poorly defined ``general'' objectives completely subverts the intention of its introduction and actually enables its use to promote\,---\,rather than counteract\,---\,claims to ``glamour and deceit'' over substantial, meaningful progress.  

\subsection{Case Studies}

We focus our analysis on two common benchmarks geared towards the assessment of general capabilities\,---\,ImageNet and GLUE, described below. The discussion is scoped to \emph{dataset}-based benchmarks, and will not include much discussion of the role of gameplay demos in AI development.

\subsubsection{ImageNet}

We characterize ImageNet as a ``general'' benchmark due to several related observations. First, at the time of development, the creators described ImageNet as representing ``the most comprehensive and diverse coverage of the image world'' \citep[p.1]{imagenet_cvpr09} and, retrospectively, Li described the project as an ``attempt to map the entire world of objects'' \citep{Gershgorn2017}. With general-purpose visual object recognition framed as the ability to recognize a sizable breadth of objects in a manner that rivals human capabilities \citep{liu2018}, we observe ImageNet's sheer size\,---\,both in terms of number of categories and number of images per category\,---\,further informs its perception as representing a general formulation of visual object recognition. At the time of its creation, it offered 20 times the number of categories, and 100 times the number of total images as the most popular training and evaluation benchmarks at the time, such as Caltech 101/256 \citep{kamarudin2015comparison, griffin2007caltech}, MSRC \citep{shotton2006textonboost} and PASCAL VOC \citep{everingham2010pascal}. Second, we observe community consensus that the task formalized in the ImageNet dataset represents a meaningful milestone towards longer-term goals of artificial visual intelligence.
Indeed, Li has explicitly characterized large scale visual object recognition as the ``north star'' of computer vision\,---\,a scientific quest that would define and guide the field towards the ultimate goal of artificial visual intelligence \citep{FeiFeiLiWhere}, with ImageNet operating as the canonical instantiation.  In their papers, ImageNet authors are clear about the desire to situate the dataset as the definitive benchmark of the field. However, the hype surrounding ImageNet and the scope of the claims derived from it are not localized the the field of computer vision. For example, increased performance on ImageNet is often explicitly referenced as an indication that the field is progressing towards general-purpose AI \citep{Sutskever2018}.

\subsection{GLUE and SuperGLUE}

The creators of GLUE (General Language Understanding Evaluation) \citep{wang2018glue} and SuperGLUE \citep{wang2019superglue} present these resources as ``evaluation framework[s] for research towards general-purpose language understanding technologies'' \citep[p.1]{wang2019superglue}, noting that, unlike human language understanding, most computer natural language understanding (NLU) systems are task or domain-specific \citep{wang2018glue}. When a human knows a language, they can use that knowledge across any task that involves that language. Thus, a benchmark that tests whether linguistic knowledge acquired through training on one task can be applied to other tasks, in principle, tests for a specific and potentially well-defined kind of generalizability.

Authors for GLUE and SuperGLUE describe both datasets as being ``designed to provide a general-purpose evaluation of language understanding that covers a range of training data volumes, task genres, and task formulations''  \citep[p.2]{wang2019superglue}. In framing these benchmarks as general language understanding evaluation benchmarks made up of a diverse set of language understanding tasks, the authors suggest the notion that success on the particular tasks included demonstrates at least progress towards a full-scale solution to language understanding, and explicitly describe such a set up as being well positioned for ``exhibiting the transfer-learning potential of approaches''  \citep[p.2]{wang2019superglue}. As a comprehensive setup for a wide range of tests in a popular language (English), this benchmark has been elevated to its current status as a general marker of progress for the NLP community, and specifically the accepted evaluation platform for ``general-purpose sentence encoders'' \citep{wang2018glue}.

\section{Limits of Benchmarking General Capabilities} 

The imagined artifact of the ``general'' benchmark does not actually exist. Real data is designed, subjective and limited in ways that necessitate a different framing from that of any claim to general knowledge or general-purpose capabilities. In fact, presenting any single dataset in this way is ultimately dangerous and deceptive, resulting in misguidance on task design and focus, under-reporting of the many biases and subjective interpretations inherent in the data as well as enabling, through false presentations of performance, potential model misuse. In this section, we walk through the key arguments for how benchmarking is a limited approach to assess general model capabilities and, in particular, discuss the risk of making this claim to generality in the context of the \emph{limited task design}, \emph{de-contextualized data and performance reporting} as well as \emph{inappropriate community use} common to such benchmarks in the ML context. For each of these limitations, we break down the shortcoming and discuss the cited evidence and reasoning for our observations. 

\subsection{Limited Task Design}
\label{sec:limit-design}

There is nothing systematic or organized about the definition and arrangement of the rooms in the museum that Grover visits. He walks through arbitrary rooms of ``Things That Make So Much Noise You Can't Hear Yourself Think'', ``The Small Hall'', ``The Carrot Room'' and the like.

Similarly, the task formation for these benchmarks seems to happen independently of the intended and declared problem space.
\citet{Schlangen2020TargetingTB} defines a \emph{task} as a mapping from an input space to an output space, defined both \textit{intensionally} (via a description of the task) and \textit{extensionally} (via a particular dataset, i.e.\ pairs of inputs and outputs matching the description). In machine learning, the tendency is towards the latter scenario, where the benchmark task is defined by a dataset, often collected with such a task in mind \citep{scheuerman2021datasets}. If the so-called ``general'' benchmarks were legitimate tests of progress towards general artificial cognitive abilities, we would expect the tasks they embody to be chosen systematically, or with reference to specific theories of the cognitive abilities they model. Instead, what we observe looks more like samples of convenience: tasks and collections of tasks arbitrarily built out of what is 
easily available to the team developing these benchmarks, even if such constructions are theoretically unsound. Unlike with software benchmarking \citep{lewis1985evolution}, the subtasks we see in machine learning ``general'' benchmarks are not axiomatic\,---\,they were not actively designed as abstractions of any meaningful general function or sub-function, and are often not systematically curated in nature. Instead, what we find in the ``general'' benchmarks are sets of categories reminiscent of the inconsistent hodgepodge of rooms that Grover visits in the ``Everything in the Whole Wide World Museum''.

\subsubsection{Arbitrarily Selected Tasks and Collections}
\label{sec:design}

Concerns of this sort were raised during early benchmark development in the field. UCI \citep{Dua:2019} was one of the first foci of shared tasks in machine learning development and contains datasets which pertain to a collection of individual subtasks, with community benchmarking practice coming to focus over time on the Iris, Adult, Wine, and Breast Cancer classification datasets.
Kiri Wagstaff voiced concern for a poor logic in the selection of this combination of subtasks. She noted that none of tasks included in this collection represent any reasonable proxy or abstraction of real-world problems even scientists in the respective fields that produced each dataset would care about. She notes, ``Legions of (ML) researchers have chased after the best iris or mushroom classifier. Yet this flurry of effort does not seem to have had any impact on the fields of botany or mycology. Do scientists in these disciplines even need such a classifier? Do they publish about this subject in their journals?'' \citep[p.2]{Wagstaff:2012}.

The classes in ImageNet are also seemingly arbitrary, ranging from specific dog breeds to high-level notions like ``New Zealand beach'' \citep{russakovskyImageNetLargeScale2015}. ImageNet labels are derived from the 12 ``subtrees'' of the WordNet ontology \citep{miller1995wordnet}: mammal, bird, fish, reptile, amphibian, vehicle, furniture, musical instrument, geological formation, tool, flower, and fruit \citep{imagenet_cvpr09}. Despite being developed in another field, with another purpose entirely, the English-language WordNet hierarchy was adopted, nearly whole cloth, for use in structuring ImageNet. Little consideration was given to the impact of the inclusion of certain words in the task of image classification. The consequences of this oversight were made tangible by the work of \citet{prabhu2020large} and \cite{crawford2019excavating} which revealed the existence of a range of derogatory and offensive categories in the ``person'' subtree. The primary categorical consideration the ImageNet creators \emph{did} discuss center around the inclusion of certain categories to ensure consistency with the categories in a previously established PASCAL benchmark \citep{russakovskyImageNetLargeScale2015}.

Similarily, the subtasks of the GLUE benchmark are not much more carefully selected. The benchmark was initially designed as a suite with a stated goal to ``spur development of generalizable NLU systems'' \citep[p.1]{wang2018glue}, embodying the idea that a system capable of using linguistic knowledge to understand language input would be able to do so across a variety of tasks. The included tasks were curated following a set of 30 proposals sourced from with an informal survey of colleagues in the NLP community \citep{wang2018glue}, and thus only really represent a collection of tasks authors from the NLP community perceived as interesting problems at the time. To filter through proposals, authors mainly rely on practical heuristics like ``licensing issues, complex formats, and insufficient headroom'' [{\it Ibid.}\ p.\ 5] before referring to high level criteria to filter down to the final eight or nine included tasks. As a result, the collection of tasks included in the final benchmark neither systematically map out a range of specific linguistic skills required for understanding (lacking in particular any exploration of pragmatics \citep{Qiu-etal2020}, or the proper handling of negation \citep{ettinger2020bert}) nor present a truly varied range of ways to deploy linguistic knowledge in comprehension. While such a collection can effectively demonstrate a model's ability to generalize performance across the included tasks, this ability is not equivalent to general language understanding, nor is it evidence that the solution to any task actually involves understanding the language in a sense that linguists might recognize. 

\subsubsection{Critical Misunderstandings of Domain Knowledge and Application Problem Space}
\label{sec:domain}

GLUE and SuperGLUE benchmarks combine linguistic competence (the ability to model a linguistic system) with general commonsense and world reasoning as if they were equivalently scoped problems. In reference to a diagnostic component of the benchmark, authors state that ``this dataset is designed to highlight common challenges, such as the use of world knowledge and logical operators, that we expect models must handle to robustly solve the tasks'' \citep[p.1]{wang2018glue}. Thus a benchmark designed to test for generalizability across different language understanding tasks comes to subsume not only the task of building up linguistic competence (e.g. logical operators) in the language in question (English) but also the ability to acquire and deploy world knowledge. It is well established that natural language understanding requires both linguistic processing and reasoning over the combination of the linguistic signal, communicative common ground, and world knowledge \citep{hunter2018formal}, but while linguistic knowledge is relatively self-contained and reusable across different textual domains, world knowledge is open-ended. Conflating these two abilities in a benchmark that is easily (mis)interpreted as representing a much more general, flexible, and robust set of capabilities than it possesses, and thus inappropriately presented as comparable to the ``human ability to understand language'' \citep{wang2018glue}.

Furthermore, language understanding relies not only on linguistic competence but also world knowledge, commonsense reasoning, and the ability to model the interlocutor's state of mind \citep{Reddy:79,Clark96}, none of which can be thoroughly tested through text-only tasks, such as GLUE.Several researchers have raised the need to establish effective physical and social grounding as part of the process of moving towards robust and effective natural language understanding, warning against text-only learning as a limited approach \citep{bisk2020experience,zellers2020evaluating}. \citet{bender-koller-2020-climbing} additionally mention the tendency of machine learning researchers to misinterpret certain benchmarks as capturing the model's ability to decipher \emph{meaning} in language, arguing that benchmarks need to be constructed with care if they are to show evidence of ``understanding'' as opposed to merely the ability to manipulate linguistic form sufficiently to pass the test. 

\subsection{De-contextualized Data and Performance Reporting}

In ``The Hall of Very, Very Light Things'', Grover finds a big rock and declares, ``There has been some mistake! This big rock is not light,'' before deciding to move it to the ``The Hall of Very, Very Heavy Things''. However, such judgements are ultimately relative\,---\,that rock is certainly much less heavy than the trailer truck in the latter room, and could be considered light in comparison. Nothing about the museum is neutrally determined.

In this section, we explore one of the features that can lead researchers to mistakenly construe a benchmark as ``general'', namely the de-contextualization of its component tasks and datasets. No dataset is neutral and there are inherent limits to what a benchmark can tell us\,---\,in fact, data benchmarks are closed and inherently subjective, localized constructions. If anything, the claim to generality will often act as cover, allowing those developing the benchmarks to escape the responsibility of reporting details of these limitations. Part of the challenge of addressing this lack of context is proper documentation for these datasets, which is often underdeveloped \citep{gebru2020datasheets,bender2018} and the devaluation of the data work \citep{paullada2020data, sambasivan2021everyone, Hutchinson2020, jo2020archives}.

\subsubsection{Limited Scope}
\label{sec:scope}

Even large datasets like ImageNet are ultimately closed systems, limited in their coverage of non-Western contexts and temporally bounded. \cite{torralba2011unbiased} demonstrate how images from the same class but different datasets are often distinguishable and embody a very specific style of capturing some segment of the real world. Specifically, a well-critiqued limitation of ImageNet is that the objects tend to be centered within the images, which does not reflect how ``natural'' images actually appear \citep{barbu2019objectnet}. Additionally, it's not clear that increasing the size of these benchmarks to capture omitted content is even practically feasible in many cases\,---\,ImageNet authors note the trade-off between annotation quality and size, remarking that ``the scale [is] already imposing limits on the manual annotations that are feasible to obtain'' \citep[p.34]{russakovskyImageNetLargeScale2015}. There are thus likely serious practical limits to making such benchmarks larger.

The notion of task diversity, not size, is repeated quite frequently by GLUE authors \citep[p.\ 1,2,3,5]{wang2018glue} as a main differentiator from predecessors' benchmarks, SentEval \citep{conneau2018senteval} and decaNLP \citep{mccann2018natural}. However, despite the expressed desire for the benchmark to include a diverse range of tasks that ``cover a diverse range of text genres, dataset sizes, and degrees of difficulty'' \citep[p.\ 1]{wang2018glue}, compared to human linguistic activity, the GLUE tasks are hardly diverse: two single-sentence tasks, two similarity and paraphrase tasks, and four inference tasks. As noted in the development of a subsequent iteration of the benchmark,  SuperGLUE, ``task formats in GLUE are limited to sentence- and sentence-pair classification''\citep[p.\ 1]{wang2019superglue}, requiring a subsequent expansion to include coreference resolution and question answering (QA) task formats in SuperGLUE. SuperGLUE includes four question answering tasks, two inference tasks, one word sense disambiguation task and one coreference task.

\subsubsection{Benchmark Subjectivity}
\label{sec:subj}

All datasets come with an embedded perspective\,---\,there is no neutral or universal dataset \citep{haraway1988situated,stitzlein2004replacing,Gebru:20, denton2021genealogy, scheuerman2021datasets}. To present a dataset, inherently both political and value-laden, as a completely neutral scientific artifact is irresponsible. In benchmarks promoted to assess general capabilities, such as ImageNet and GLUE, such politics remain unacknowledged\,---\,undiscussed and hidden for the sake of maintaining the claim to broad relevance. However, denying DRdelete{the existence of unacknowledged} context does not make it disappear. In fact, this distorted data lens is often not limited in an arbitrary way, but limited in a way that hurts certain groups of people\,---\,those without the power to define the data themselves. For example, people who hold transgender and gender-nonconforming gender identities and non-white, non-Western racial identities are underrepresented in mainstream face datasets \citep{merler2019diversity}, and images of members of these communities are often tagged with racial or ethnic slurs, even on large general use datasets such as MIT Tiny Images or ImageNet \citep{crawford2019excavating,gehl2017training,prabhu2020large}. 

GLUE and SuperGLUE target one specific language (i.e. English), not ``language'' in the abstract. This very specific subset of American English text in the presence of annotation artifacts \citep{gururangan2018annotation} results in inherently subjective outcomes \citep{waseem2020disembodied}. \footnote{\citet{waseem2020disembodied} are discussing NLP datasets and modeling in general, not GLUE, SuperGLUE or other similar benchmarks specifically. However, their general points apply: any given dataset represents the embodied viewpoint of its authors and annotators, and furthermore, datasets constructed without attention to whose viewpoints are being represented will likely over-represent hegemonic ones. See also \citet{Ben:Geb:McM:21}.} 

The ImageNet creators did attempt to diversify their dataset by translating image queries into other languages, including Chinese, Spanish, Dutch and Italian to source representations for the final English-labeled categories \citep{imagenet_cvpr09}. However, when analyzing country level geo-location data for the 2011 version of the dataset, 45\% of the images were found to be sourced from the US, and over 60\% from a small selection of Western countries in North America and Europe \citep{shankar2017no}. In fact, only 1\% and 1.2\% of images are from China and India respectively, despite the fact that those countries are the most populous countries on the planet \citep{shankar2017no}. This lack of geo-diversity of sources manifests in poor dataset representation, as the dataset becomes visually anchored to a specific dominant cultural context. \citet{de2019does} found that if ImageNet was developed using the results of Hindi online image queries, rather than English ones, the result would be a dataset that looks drastically different, embodying completely new visual representations of not only socially defined concepts, such as ``wedding'', but also everyday objects, such as ``spice''. 

\subsection{Inappropriate Community Use}

When Grover arrived at the museum, he was excited and inspired by its claim to contain every object in the ``Whole Wide World''\,---\,however, once he enters, his focus is only redirected to whatever is selected to be highlighted in the museum. Despite his initial interest being motivated by the pursuit of generality, it is only really the subset of consciously-included objects that actually captures his attention.

When individual dataset creators and communities overstate the generality of these benchmark datasets, they elevate them to the status of a target the entire field should be aiming for. This benchmark focus can escalate to the point of researchers falling into the trap of uncritically chasing algorithmic improvement as measured by these datasets, losing sight of the potential for performance mismatch with the real world or more relevant problem formulations. 

\subsubsection{Limits of Competitive Testing}
\label{sec:competition}

Chasing ``state of the art'' (SOTA) performance is a very peculiar way of doing science\,---\,one that focuses on empirical and incremental work rather than hypothesis-based scientific inquiry \citep{hooker1995testing}. In 1995, Lorenza Saitta criticized benchmark chasing as ``allow[ing] researchers to publish dull papers that proposed small variations of existing supervised learning algorithms and reported their small-but-significant incremental performance improvements in comparison studies'' \citep[p.\ 61]{radinDigitalNativesHow2017}. Thomas and Uminsky go so far as to present metric chasing as an ethical issue, stating that ``overemphasizing metrics leads to manipulation, gaming, a myopic focus on short-term goals, and other unexpected negative consequences'' \citep[p.\ 1]{thomas2020problem}. 

Furthermore, Kiri Wagstaff points out some inherent flaws of the aggregate performance evaluation format for many ``general'' benchmarks and leader boards. She notes that ``80\% accuracy on Iris classification might be sufficient for the botany world, but to classify as poisonous or edible a mushroom you intend to ingest, perhaps 99\% (or higher) accuracy is required. The assumption of cross-domain comparability is a mirage created by the application of metrics that have the same range, but not the same meaning. Suites of experiments are often summarized by the average accuracy across all datasets. This tells us nothing at all useful about generalization or impact, since the meaning of an x\% improvement may be very different for different data sets (or classes).'' \citep[p.3]{Wagstaff:2012} Wagstaff here calls out the fact that the task is not just poorly formulated but also likely measured inappropriately, in a way that makes the results difficult to interpret meaningfully.

Despite these issues, both GLUE and ImageNet authors situate the benchmark datasets within designed competitive environments. GLUE authors cite ``the leaderboard and error analysis toolkit'' as a differentiating factor \citep[p.3]{wang2018glue}, while ImageNet authors detail the format of ``an annual competition and corresponding workshop''  \citep[p.1,49]{russakovskyImageNetLargeScale2015}.

\subsubsection{Redirection of Focus for the Field}
\label{sec:focus}

Benchmarks have always been historically influential, and are often created intentionally to incentivize interest in certain research topics. The National Institute of Standards and Technology (NIST) invested in excess of \$6.5 million for the FERET facial recognition challenge from 1993 to 1998, in order to encourage research participation on a technology of interest to the sponsoring agency\,---\,the Department of Defense \citep{phillips2000feret}. The Netflix Prize, which ran from 2006 to 2009, was launched by Netflix, an online DVD-rental and video streaming service, in order to drive innovation for the development of collaborative filtering algorithms that improved the quality of automated movie recommendations \citep{bennett2007netflix}. However, there is a notable difference between such competitions and those involving what are presented as ``general'' benchmarks\,---\,namely, the PASCAL VOC challenge running from 2005 to 2012, the ImageNet Large Scale Visual Recognition Challenge (ILSVRC) running annually from 2010 to 2017, and the GLUE/SuperGLUE leaderboard, which has been running since 2018. The focus of competitions involving ``general'' benchmarks are much further detached from grounded applications, with tasks more vaguely defined. As a result, these competitions become interpreted as significant markers of success across a variety of relevant sub-domains, and their popularity grows accordingly as that competition is elevated to represent the ultimate showcase of algorithmic performance. 

Benchmarks also influence the nature of the dominant algorithmic approaches attempted. In the 1960s, chess caused the AI community to hyper-focus on deep-tree searching and the minimax algorithms, which were most effective on improving game performance. Both of these methods came to dominate the algorithmic development of this time, resulting in the neglect of alternate problems and approaches \citep{Ensmenger:12}. \citet{dotan2019valueladen} note how trends for algorithmic development in ML are driven by performance on particular benchmarks, and how these trends result in shifts to the entire research program of the field.  ``General'' benchmarks in particular often inspire resource-intensive models: both through presenting very large training sets for models to consume and by conflating size (of both training set and thus model) with generality. The current deep learning movement would not have been possible without ImageNet \citep{alom2018history}. In a similar way, the success of BERT models, a breakthrough of language model development, was first effectively demonstrated using GLUE \citep{devlin2018bert}.

\subsubsection{Justification for Practical out of Context or Unsafe Applications}
\label{sec:unsafe}

General benchmarks also get mentioned in marketing copy for commercial machine learning products, with performance on the benchmark presented as evidence of real-world technical achievement. This context is when the significance of benchmarks is most severely distorted, when performance on benchmarks is not just the tool for algorithmic selection, but actually presented as some reliable marker of expected model achievement in deployment. For instance, in January 2021, Microsoft claimed ``DeBERTa surpassing human performance on SuperGLUE marks an important milestone toward general AI.'' \citep{MSRGlue}.ImageNet is equally elevated as an indicator of commercial model success\,---\,so much so that Baidu cheated on the benchmark in order to inflate product performance claims in marketing \citep{simonite2015and}.

Poor performance on minority subgroups could possibly be disguised in aggregate and de-contextualized ``general'' benchmarks to justify deployment. We already know from examples of deployed, commercial products in facial recognition \citep{GenderShades} that unbalanced representation impacts performance on certain groups over others, though this disproportionate performance is often hidden in aggregate performance measures on what turn out to be biased benchmarks \citep{merler2019diversity, Raji2019}. 
 
\section{Alternative Roles for Benchmarking and Alternative Evaluation Methods}

In preceding sections, we have explored the ways in which ``general'' benchmarks fail to serve as effective measures of progress in machine learning. Now we ask: What can be done instead? This is not a question of ``fixing'', ``improving'' or ``expanding'' the current ``general'' benchmarks\,---\,after all, the solution to addressing the issues inherent to the ``Everything in the Whole Wide World'' museum is not to add more rooms. The presentation of a data benchmark as being able to exist independent of context, scope and specificity is itself a false premise for machine learning evaluation.

Instead, we note two paths forward. First, we argue that benchmarks should be developed, presented and understood as intended\,---\,to evaluate concrete, well-scoped and contextualized tasks. Second, when probing for more broad model objectives, behaviors and capabilities, we propose that the field explore alternative evaluation methods. The following are some starting points on evaluation techniques to explore further in machine learning as an alternative to benchmarking (further details in Appendix~\ref{appa}):

\begin{itemize}
\item \emph{The systematic development of test items} (see Appendix \ref{sec:ts}), in the form of testsuites \citep[e.g.][]{TSNLP,ribeiro-etal-2020-beyond}, audits \citep[e.g.][]{GenderShades}, and adversarial testing \citep[e.g.][]{ettinger-etal-2017-towards,Niv:Kao:19}. These techniques can help map out which aspects of the problem space remain challenging and/or check for the potential for harms coming from system biases. 
\item \emph{System output analysis / behavioural testing} (see Appendix \ref{sec:output}), including error analysis \citep[e.g.][]{van:Cli:Gka:21}, disaggregated analysis \citep[e.g.][]{Raji2019}, and counterfactual analysis \citep[e.g.][]{Garg2019, Hutchinson2020}. These techniques can reveal the kinds of failure modes a system might produce, and potentially their causes. 
\item \emph{Ablation testing} (see Appendix \ref{sec:ablation}) \citep[e.g.][]{Newell:75} allows researchers insight into the specific contributions of different system components. 
\item \emph{Techniques for analyzing model properties} (see Appendix \ref{sec:mp}) that are orthogonal to system outputs such as profiling energy consumption \citep[e.g.][]{Henderson:2020, Schwartz2019}, memory requirements \citep[e.g.][]{Ethayarajh2020}, and stability in the face of perturbations to training data \citep[e.g.][]{Sculley2018WinnersCO}.

\end{itemize}

\section{Conclusion}

The situation with Grover and the museum's claims are clearly ridiculous\,---\,yet in machine learning, we follow the exact same logical fallacies to justify the elevation of a select number of benchmarks operating as general benchmarks for the field. However, there is no dataset that will be able to capture the full complexity of the details of existence, in the same way that there can be no museum to contain the full catalog of everything in the whole wide world. Open-world, universal and neutral datasets don't exist, and current methods of benchmarking do not offer meaningful measures of general capabilities.

The effective development of benchmarks is critical to progress in machine learning, but what makes a benchmark effective is not the strength of its arbitrary and false claim to ``generality'' but its effectiveness in how it  helps us understand as researchers how certain systems work\,---\,and how they don't. Benchmarking, appropriately deployed, is not about winning a contest but more about surveying a landscape\,---\,the more we can re-frame, contextualize and appropriately scope these datasets, the more useful they will become as an informative dimension to more impactful algorithmic development and alternative evaluation methods (see Appendix \ref{appa}). Given the alternative roles and interpretations for evaluation we could explore, it is essential that we move quickly beyond the narrow-yet-totalizing lens of the ``Everything in the Whole Wide World'' benchmarks.

\section{Acknowledgments}
The authors thank Sam Bowman, Julian Michael, and Ludwig Schmidt for comments on earlier drafts of this work, as well as the organizers and attendees of the NeurIPS 2020 Workshop on ML Retrospectives, Surveys, and Meta-Analyses for providing an early venue for this work. We also thank the reviewers for their comments.

\bibliographystyle{plainnat}
\bibliography{bibliography}

\begin{thebibliography}{105}
\providecommand{\natexlab}[1]{#1}
\providecommand{\url}[1]{\texttt{#1}}
\expandafter\ifx\csname urlstyle\endcsname\relax
  \providecommand{\doi}[1]{doi: #1}\else
  \providecommand{\doi}{doi: \begingroup \urlstyle{rm}\Url}\fi

\bibitem[Adam(1998)]{Adam98}
Alison Adam.
\newblock \emph{Artificial Knowing: Gender and the Thinking Machine}.
\newblock Routledge, USA, 1998.
\newblock ISBN 0415129621.

\bibitem[Alom et~al.(2018)Alom, Taha, Yakopcic, Westberg, Sidike, Nasrin,
  Van~Esesn, Awwal, and Asari]{alom2018history}
Md~Zahangir Alom, Tarek~M Taha, Christopher Yakopcic, Stefan Westberg, Paheding
  Sidike, Mst~Shamima Nasrin, Brian~C Van~Esesn, Abdul A~S Awwal, and Vijayan~K
  Asari.
\newblock The history began from {AlexNet}: {A} comprehensive survey on deep
  learning approaches.
\newblock \emph{arXiv preprint arXiv:1803.01164}, 2018.

\bibitem[Barbu et~al.(2019)Barbu, Mayo, Alverio, Luo, Wang, Gutfreund,
  Tenenbaum, and Katz]{barbu2019objectnet}
Andrei Barbu, David Mayo, Julian Alverio, William Luo, Christopher Wang, Dan
  Gutfreund, Josh Tenenbaum, and Boris Katz.
\newblock Objectnet: A large-scale bias-controlled dataset for pushing the
  limits of object recognition models.
\newblock In \emph{Advances in Neural Information Processing Systems}, pages
  9453--9463, 2019.

\bibitem[Belz and Kilgarriff(2006)]{belz2006shared}
Anja Belz and Adam Kilgarriff.
\newblock Shared-task evaluations in {{HLT}}: {{Lessons}} for {{NLG}}.
\newblock In \emph{Proceedings of the Fourth International Natural Language
  Generation Conference}, pages 133--135, 2006.

\bibitem[Bender and Friedman(2018)]{bender2018}
Emily~M. Bender and Batya Friedman.
\newblock Data statements for natural language processing: Toward mitigating
  system bias and enabling better science.
\newblock \emph{Transactions of the Association for Computational Linguistics},
  6:\penalty0 587--604, 2018.
\newblock \doi{10.1162/tacl_a_00041}.
\newblock URL \url{https://www.aclweb.org/anthology/Q18-1041}.

\bibitem[Bender and Koller(2020)]{bender-koller-2020-climbing}
Emily~M. Bender and Alexander Koller.
\newblock Climbing towards {NLU}: {On} meaning, form, and understanding in the
  age of data.
\newblock In \emph{Proceedings of the 58th Annual Meeting of the Association
  for Computational Linguistics}, pages 5185--5198, Online, July 2020.
  Association for Computational Linguistics.
\newblock \doi{10.18653/v1/2020.acl-main.463}.
\newblock URL \url{https://www.aclweb.org/anthology/2020.acl-main.463}.

\bibitem[Bender et~al.(2021)Bender, Gebru, McMillan-Major, and
  Shmitchell]{Ben:Geb:McM:21}
Emily~M. Bender, Timnit Gebru, Angelina McMillan-Major, and Shmargaret
  Shmitchell.
\newblock On the dangers of stochastic parrots: Can language models be too
  big?\raisebox{-5pt}{\includegraphics[scale=0.075]{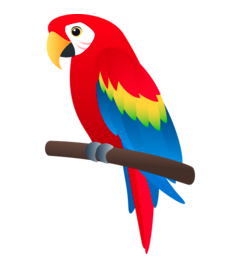}}.
\newblock In \emph{Proceedings of FAccT 2021}, 2021.

\bibitem[Bennett et~al.(2007)Bennett, Lanning, et~al.]{bennett2007netflix}
James Bennett, Stan Lanning, et~al.
\newblock The {Netflix} prize.
\newblock In \emph{Proceedings of KDD cup and workshop}, volume 2007, page~35.
  New York, 2007.

\bibitem[Birhane and Prabhu(2021)]{prabhu2020large}
Abeba Birhane and Vinay~Uday Prabhu.
\newblock Large image datasets: {A} pyrrhic win for computer vision?
\newblock In \emph{2021 {IEEE} Winter Conference on Applications of Computer
  Vision (WACV)}, pages 1536--1546, 2021.
\newblock \doi{10.1109/WACV48630.2021.00158}.

\bibitem[Bisk et~al.(2020)Bisk, Holtzman, Thomason, Andreas, Bengio, Chai,
  Lapata, Lazaridou, May, Nisnevich, Pinto, and Turian]{bisk2020experience}
Yonatan Bisk, Ari Holtzman, Jesse Thomason, Jacob Andreas, Yoshua Bengio, Joyce
  Chai, Mirella Lapata, Angeliki Lazaridou, Jonathan May, Aleksandr Nisnevich,
  Nicolas Pinto, and Joseph Turian.
\newblock Experience grounds language.
\newblock In \emph{Proceedings of the 2020 Conference on Empirical Methods in
  Natural Language Processing (EMNLP)}, pages 8718--8735, Online, November
  2020. Association for Computational Linguistics.
\newblock URL \url{https://www.aclweb.org/anthology/2020.emnlp-main.703}.

\bibitem[Bommasani et~al.(2021)Bommasani, Hudson, Adeli, Altman, Arora, von
  Arx, Bernstein, Bohg, Bosselut, Brunskill,
  et~al.]{bommasani2021opportunities}
Rishi Bommasani, Drew~A Hudson, Ehsan Adeli, Russ Altman, Simran Arora, Sydney
  von Arx, Michael~S Bernstein, Jeannette Bohg, Antoine Bosselut, Emma
  Brunskill, et~al.
\newblock On the opportunities and risks of foundation models.
\newblock \emph{arXiv preprint arXiv:2108.07258}, 2021.

\bibitem[Bowman and Dahl(2021)]{bowman2021will}
Samuel~R. Bowman and George Dahl.
\newblock What will it take to fix benchmarking in natural language
  understanding?
\newblock In \emph{Proceedings of the 2021 Conference of the North American
  Chapter of the Association for Computational Linguistics: Human Language
  Technologies}, pages 4843--4855, Online, June 2021. Association for
  Computational Linguistics.
\newblock \doi{10.18653/v1/2021.naacl-main.385}.
\newblock URL \url{https://aclanthology.org/2021.naacl-main.385}.

\bibitem[Buolamwini and Gebru(2018)]{GenderShades}
Joy Buolamwini and Timnit Gebru.
\newblock Gender shades: {Intersectional} accuracy disparities in commercial
  gender classification.
\newblock In \emph{Conference on Fairness, Accountability and Transparency},
  pages 77--91, 2018.

\bibitem[Christensen and Connault(2019)]{Christensen2019}
Timothy Christensen and Benjamin Connault.
\newblock Counterfactual sensitivity and robustness, 2019.
\newblock https://arxiv.org/abs/1904.00989.

\bibitem[Church(2018)]{churchEmergingTrendsTribute2018}
Kenneth~Ward Church.
\newblock Emerging trends: {{A}} tribute to {{Charles Wayne}}.
\newblock \emph{Natural Language Engineering}, 24\penalty0 (1):\penalty0
  155--160, January 2018.
\newblock ISSN 1351-3249, 1469-8110.
\newblock \doi{10.1017/S1351324917000389}.

\bibitem[Clark(1996)]{Clark96}
Herbert~H. Clark.
\newblock \emph{Using Language}.
\newblock Cambridge University Press, Cambridge, 1996.

\bibitem[Conneau and Kiela(2018)]{conneau2018senteval}
Alexis Conneau and Douwe Kiela.
\newblock Senteval: An evaluation toolkit for universal sentence
  representations.
\newblock \emph{arXiv preprint arXiv:1803.05449}, 2018.

\bibitem[Crawford and Paglen(2019)]{crawford2019excavating}
Kate Crawford and Trevor Paglen.
\newblock Excavating {AI}: {The} politics of images in machine learning
  training sets.
\newblock \emph{Excavating AI}, 2019.

\bibitem[de~Vries et~al.(2019)de~Vries, Misra, Wang, and van~der
  Maaten]{de2019does}
Terrance de~Vries, Ishan Misra, Changhan Wang, and Laurens van~der Maaten.
\newblock Does object recognition work for everyone?
\newblock In \emph{Proceedings of the {IEEE} Conference on Computer Vision and
  Pattern Recognition Workshops}, pages 52--59, 2019.

\bibitem[Deng et~al.(2009)Deng, Dong, Socher, Li, Li, and
  Fei-Fei]{imagenet_cvpr09}
Jia Deng, Wei Dong, Richard Socher, Li-Jia Li, Kai Li, and Li~Fei-Fei.
\newblock Imagenet: A large-scale hierarchical image database.
\newblock In \emph{2009 IEEE conference on computer vision and pattern
  recognition}, pages 248--255. Ieee, 2009.

\bibitem[Denton et~al.(2021)Denton, Hanna, Amironesei, Smart, and
  Nicole]{denton2021genealogy}
Emily Denton, Alex Hanna, Razvan Amironesei, Andrew Smart, and Hilary Nicole.
\newblock On the genealogy of machine learning datasets: A critical history of
  imagenet.
\newblock \emph{Big Data \& Society}, 8\penalty0 (2):\penalty0
  20539517211035955, 2021.
\newblock \doi{10.1177/20539517211035955}.
\newblock URL \url{https://doi.org/10.1177/20539517211035955}.

\bibitem[Devlin et~al.(2019)Devlin, Chang, Lee, and Toutanova]{devlin2018bert}
Jacob Devlin, Ming-Wei Chang, Kenton Lee, and Kristina Toutanova.
\newblock {BERT}: {Pre-training} of deep bidirectional transformers for
  language understanding.
\newblock In \emph{Proceedings of the 2019 Conference of the North {A}merican
  Chapter of the Association for Computational Linguistics: Human Language
  Technologies, Volume 1 (Long and Short Papers)}, pages 4171--4186,
  Minneapolis, Minnesota, June 2019. Association for Computational Linguistics.
\newblock \doi{10.18653/v1/N19-1423}.
\newblock URL \url{https://aclanthology.org/N19-1423}.

\bibitem[Dodge et~al.(2019)Dodge, Gururangan, Card, Schwartz, and
  Smith]{dodge2019}
Jesse Dodge, Suchin Gururangan, Dallas Card, Roy Schwartz, and Noah~A. Smith.
\newblock Show your work: Improved reporting of experimental results.
\newblock In \emph{Proceedings of the 2019 Conference on Empirical Methods in
  Natural Language Processing and the 9th International Joint Conference on
  Natural Language Processing (EMNLP-IJCNLP)}, pages 2185--2194, Hong Kong,
  China, November 2019. Association for Computational Linguistics.
\newblock \doi{10.18653/v1/D19-1224}.
\newblock URL \url{https://www.aclweb.org/anthology/D19-1224}.

\bibitem[Donoho(2017)]{donoho50YearsData2017}
David Donoho.
\newblock 50 {{Years}} of {{Data Science}}.
\newblock \emph{Journal of Computational and Graphical Statistics}, 26\penalty0
  (4):\penalty0 745--766, October 2017.
\newblock ISSN 1061-8600.
\newblock \doi{10.1080/10618600.2017.1384734}.

\bibitem[Dotan and Milli(2020)]{dotan2019valueladen}
Ravit Dotan and Smitha Milli.
\newblock Value-laden disciplinary shifts in machine learning.
\newblock In \emph{Proceedings of the 2020 Conference on Fairness,
  Accountability, and Transparency}, FAT* '20, page 294, New York, NY, USA,
  2020. Association for Computing Machinery.
\newblock ISBN 9781450369367.
\newblock \doi{10.1145/3351095.3373157}.
\newblock URL \url{https://doi.org/10.1145/3351095.3373157}.

\bibitem[Du et~al.(2020)Du, Ott, Li, Zhou, and Stoyanov]{du2020}
Jingfei Du, Myle Ott, Haoran Li, Xing Zhou, and Veselin Stoyanov.
\newblock General purpose text embeddings from pre-trained language models for
  scalable inference.
\newblock In \emph{Findings of the Association for Computational Linguistics:
  EMNLP 2020}, pages 3018--3030, Online, November 2020. Association for
  Computational Linguistics.
\newblock URL \url{https://www.aclweb.org/anthology/2020.findings-emnlp.271}.

\bibitem[Dua and Graff(2017)]{Dua:2019}
Dheeru Dua and Casey Graff.
\newblock {UCI} machine learning repository, 2017.
\newblock http://archive.ics.uci.edu/ml.

\bibitem[Ensmenger(2012)]{Ensmenger:12}
Nathan Ensmenger.
\newblock Is chess the drosophila of artificial intelligence? {A} social
  history of an algorithm.
\newblock \emph{Social Studies of Science}, 42\penalty0 (1):\penalty0 5--30,
  2012.
\newblock \doi{10.1177/0306312711424596}.
\newblock URL \url{https://doi.org/10.1177/0306312711424596}.
\newblock PMID: 22530382.

\bibitem[Ethayarajh and Jurafsky(2020)]{Ethayarajh2020}
Kawin Ethayarajh and Dan Jurafsky.
\newblock Utility is in the eye of the user: A critique of {NLP} leaderboard
  design.
\newblock In \emph{Proceedings of the 2020 Conference on Empirical Methods in
  Natural Language Processing (EMNLP)}, pages 4846--4853, Online, November
  2020. Association for Computational Linguistics.
\newblock URL \url{https://www.aclweb.org/anthology/2020.emnlp-main.393}.

\bibitem[Ettinger(2020)]{ettinger2020bert}
Allyson Ettinger.
\newblock What {BERT} is not: {Lessons} from a new suite of psycholinguistic
  diagnostics for language models.
\newblock \emph{Transactions of the Association for Computational Linguistics},
  8:\penalty0 34--48, 2020.

\bibitem[Ettinger et~al.(2017)Ettinger, Rao, Daum{\'e}~III, and
  Bender]{ettinger-etal-2017-towards}
Allyson Ettinger, Sudha Rao, Hal Daum{\'e}~III, and Emily~M. Bender.
\newblock Towards linguistically generalizable {NLP} systems: A workshop and
  shared task.
\newblock In \emph{Proceedings of the First Workshop on Building Linguistically
  Generalizable {NLP} Systems}, pages 1--10, Copenhagen, Denmark, September
  2017. Association for Computational Linguistics.
\newblock \doi{10.18653/v1/W17-5401}.
\newblock URL \url{https://www.aclweb.org/anthology/W17-5401}.

\bibitem[Everingham et~al.(2010)Everingham, Van~Gool, Williams, Winn, and
  Zisserman]{everingham2010pascal}
Mark Everingham, Luc Van~Gool, Christopher~KI Williams, John Winn, and Andrew
  Zisserman.
\newblock The {{\sc Pascal}} visual object classes ({VOC}) challenge.
\newblock \emph{International Journal of Computer Vision}, 88\penalty0
  (2):\penalty0 303--338, 2010.

\bibitem[Fancellu and Webber(2015)]{fancellu-webber-2015-translating-negation}
Federico Fancellu and Bonnie Webber.
\newblock Translating negation: A manual error analysis.
\newblock In \emph{Proceedings of the Second Workshop on Extra-Propositional
  Aspects of Meaning in Computational Semantics ({E}x{P}ro{M} 2015)}, pages
  2--11, Denver, Colorado, June 2015. Association for Computational
  Linguistics.
\newblock \doi{10.3115/v1/W15-1301}.
\newblock URL \url{https://www.aclweb.org/anthology/W15-1301}.

\bibitem[Garg et~al.(2019)Garg, Perot, Limtiaco, Taly, Chi, and
  Beutel]{Garg2019}
Sahaj Garg, Vincent Perot, Nicole Limtiaco, Ankur Taly, Ed~H. Chi, and Alex
  Beutel.
\newblock Counterfactual fairness in text classification through robustness.
\newblock In \emph{Proceedings of the 2019 {AAAI/ACM} Conference on {AI},
  Ethics, and Society}, AIES '19, page 219–226, New York, NY, USA, 2019.
  Association for Computing Machinery.
\newblock ISBN 9781450363242.
\newblock \doi{10.1145/3306618.3317950}.
\newblock URL \url{https://doi.org/10.1145/3306618.3317950}.

\bibitem[Gebru(2020)]{Gebru:20}
Timnit Gebru.
\newblock Race and gender.
\newblock In Markus~D. Dubber, Frank Pasquale, and Sunit Das, editors,
  \emph{The {Oxford} Handbook of Ethics of {AI}}, pages 253--270. Oxford
  University Press, Oxford, 2020.

\bibitem[Gebru et~al.(2020)Gebru, Morgenstern, Vecchione, Vaughan, Wallach,
  III, and Crawford]{gebru2020datasheets}
Timnit Gebru, Jamie Morgenstern, Briana Vecchione, Jennifer~Wortman Vaughan,
  Hanna Wallach, Hal~Daumé III, and Kate Crawford.
\newblock Datasheets for datasets, 2020.
\newblock https://arxiv.org/abs/1803.09010.

\bibitem[Gehl et~al.(2017)Gehl, Moyer-Horner, and Yeo]{gehl2017training}
Robert~W Gehl, Lucas Moyer-Horner, and Sara~K Yeo.
\newblock Training computers to see internet pornography: Gender and sexual
  discrimination in computer vision science.
\newblock \emph{Television \& New Media}, 18\penalty0 (6):\penalty0 529--547,
  2017.

\bibitem[Gershgorn(2017)]{Gershgorn2017}
Dave Gershgorn.
\newblock The data that transformed {AI} research—--and possibly the world.
\newblock \emph{Quartz}, 2017.
\newblock
  https://qz.com/1034972/the-data-that-changed-the-direction-of-ai-research-and-possibly-the-world/.

\bibitem[Griffin et~al.(2007)Griffin, Holub, and Perona]{griffin2007caltech}
Gregory Griffin, Alex Holub, and Pietro Perona.
\newblock Caltech-256 object category dataset, 2007.
\newblock California Institute of Technology.

\bibitem[Grishman and Sundheim(1996)]{grishman1996}
Ralph Grishman and Beth Sundheim.
\newblock Message understanding conference-6: {{A}} brief history.
\newblock In \emph{Proceedings of the 16th Conference on Computational
  Linguistics - Volume 1}, {{COLING}} '96, pages 466--471, {USA}, 1996.
  {Association for Computational Linguistics}.
\newblock \doi{10.3115/992628.992709}.

\bibitem[Gururangan et~al.(2018)Gururangan, Swayamdipta, Levy, Schwartz,
  Bowman, and Smith]{gururangan2018annotation}
Suchin Gururangan, Swabha Swayamdipta, Omer Levy, Roy Schwartz, Samuel Bowman,
  and Noah~A. Smith.
\newblock Annotation artifacts in natural language inference data.
\newblock In \emph{Proceedings of the 2018 Conference of the North {A}merican
  Chapter of the Association for Computational Linguistics: Human Language
  Technologies, Volume 2 (Short Papers)}, pages 107--112, New Orleans,
  Louisiana, June 2018. Association for Computational Linguistics.
\newblock \doi{10.18653/v1/N18-2017}.
\newblock URL \url{https://www.aclweb.org/anthology/N18-2017}.

\bibitem[Haraway(1988)]{haraway1988situated}
Donna Haraway.
\newblock Situated knowledges: {The} science question in feminism and the
  privilege of partial perspective.
\newblock \emph{Feminist Studies}, 14\penalty0 (3):\penalty0 575--599, 1988.

\bibitem[He et~al.(2021)He, Liu, Gao, and Chen]{MSRGlue}
Pengcheng He, Xiaodong Liu, Jianfeng Gao, and Weizhu Chen.
\newblock Microsoft deberta surpasses human performance on the superglue
  benchmark, 2021.
\newblock URL
  \url{https://www.microsoft.com/en-us/research/blog/microsoft-deberta-surpasses-human-performance-on-the-superglue-benchmark/}.

\bibitem[Heinzerling(2019)]{Heinzerling:19}
Benjamin Heinzerling.
\newblock {NLP}'s {Clever} {Hans} moment has arrived.
\newblock Blog post, available at
  \url{https://bheinzerling.github.io/post/clever-hans/}, accessed July 25,
  2019, 2019.

\bibitem[Henderson et~al.(2020)Henderson, Hu, Romoff, Brunskill, Jurafsky, and
  Pineau]{Henderson:2020}
Peter Henderson, Jieru Hu, Joshua Romoff, Emma Brunskill, Dan Jurafsky, and
  Joelle Pineau.
\newblock Towards the systematic reporting of the energy and carbon footprints
  of machine learning.
\newblock \emph{Journal of Machine Learning Research}, 21\penalty0
  (248):\penalty0 1--43, 2020.
\newblock URL \url{http://jmlr.org/papers/v21/20-312.html}.

\bibitem[Hooker(1995)]{hooker1995testing}
John~N Hooker.
\newblock Testing heuristics: We have it all wrong.
\newblock \emph{Journal of heuristics}, 1\penalty0 (1):\penalty0 33--42, 1995.

\bibitem[Hossain et~al.(2020)Hossain, Anastasopoulos, Blanco, and
  Palmer]{hossain2020its}
Md~Mosharaf Hossain, Antonios Anastasopoulos, Eduardo Blanco, and Alexis
  Palmer.
\newblock It{'}s not a non-issue: Negation as a source of error in machine
  translation.
\newblock In \emph{Findings of the Association for Computational Linguistics:
  EMNLP 2020}, pages 3869--3885, Online, November 2020. Association for
  Computational Linguistics.
\newblock URL \url{https://www.aclweb.org/anthology/2020.findings-emnlp.345}.

\bibitem[Huh et~al.(2016)Huh, Agrawal, and Efros]{huh2016what}
Mi-Young Huh, Pulkit Agrawal, and A.~Alexei Efros.
\newblock What makes imagenet good for transfer learning?
\newblock \emph{NIPS Workshop on Large Scale Computer Vision Systems}, 2016.

\bibitem[Hunter et~al.(2018)Hunter, Asher, and Lascarides]{hunter2018formal}
Julie Hunter, Nicholas Asher, and Alex Lascarides.
\newblock A formal semantics for situated conversation.
\newblock \emph{Semantics and Pragmatics}, 11:\penalty0 10, 2018.

\bibitem[Hutchinson et~al.(2020)Hutchinson, Prabhakaran, Denton, Webster,
  Zhong, and Denuyl]{Hutchinson2020}
Ben Hutchinson, Vinodkumar Prabhakaran, Emily Denton, Kellie Webster, Yu~Zhong,
  and Stephen Denuyl.
\newblock Social biases in {NLP} models as barriers for persons with
  disabilities.
\newblock In \emph{Proceedings of the 58th Annual Meeting of the Association
  for Computational Linguistics}, pages 5491--5501, Online, July 2020.
  Association for Computational Linguistics.
\newblock \doi{10.18653/v1/2020.acl-main.487}.
\newblock URL \url{https://www.aclweb.org/anthology/2020.acl-main.487}.

\bibitem[Jacobs and Wallach(2021)]{Jac:Wal:21}
Abigail~Z. Jacobs and Hanna Wallach.
\newblock Measurement and fairness.
\newblock In \emph{Proceedings of the 2021 ACM Conference on Fairness,
  Accountability, and Transparency}, FAccT '21, page 375–385, New York, NY,
  USA, 2021. Association for Computing Machinery.
\newblock ISBN 9781450383097.
\newblock \doi{10.1145/3442188.3445901}.
\newblock URL \url{https://doi.org/10.1145/3442188.3445901}.

\bibitem[Jo and Gebru(2020)]{jo2020archives}
Eun~Seo Jo and Timnit Gebru.
\newblock Lessons from archives: Strategies for collecting sociocultural data
  in machine learning.
\newblock In \emph{Proceedings of the 2020 Conference on Fairness,
  Accountability, and Transparency}, FAT* '20, page 306–316, New York, NY,
  USA, 2020. Association for Computing Machinery.
\newblock ISBN 9781450369367.
\newblock \doi{10.1145/3351095.3372829}.
\newblock URL \url{https://doi.org/10.1145/3351095.3372829}.

\bibitem[Kahmen and Faig(1988)]{kahmen1988surveying}
Heribert Kahmen and Wolfgang Faig.
\newblock \emph{Surveying}.
\newblock Walter de Gruyter, Berlin, 1988.

\bibitem[Kamarudin et~al.(2015)Kamarudin, Makhtar, Fadzli, Mohamad, Mohamad,
  and Kadir]{kamarudin2015comparison}
Nur~Shazwani Kamarudin, Mokhairi Makhtar, Syed~Abdullah Fadzli, Mumtazimah
  Mohamad, Fatma~Susilawati Mohamad, and Mohd Fadzil~Abdul Kadir.
\newblock Comparison of image classification techniques using {Caltech} 101
  dataset.
\newblock \emph{Journal of Theoretical \& Applied Information Technology},
  71\penalty0 (1), 2015.

\bibitem[Kilgarri(1998)]{kilgarri1998senseval}
Adam Kilgarri.
\newblock Senseval: An exercise in evaluating word sense disambiguation
  programs.
\newblock In \emph{Proc. of the first international conference on language
  resources and evaluation}, pages 581--588, 1998.

\bibitem[Lehmann et~al.(1996)Lehmann, Oepen, Regnier-Prost, Netter, Lux, Klein,
  Falkedal, Fouvry, Estival, Dauphin, Compagnion, Baur, Balkan, and
  Arnold]{TSNLP}
Sabine Lehmann, Stephan Oepen, Sylvie Regnier-Prost, Klaus Netter, Veronika
  Lux, Judith Klein, Kirsten Falkedal, Frederik Fouvry, Dominique Estival, Eva
  Dauphin, Herv{\'e} Compagnion, Judith Baur, Lorna Balkan, and Doug Arnold.
\newblock {\sc tsnlp} --- {T}est {S}uites for {N}atural {L}anguage
  {P}rocessing.
\newblock In \emph{Proceedings of the 16th {I}nternational {C}onference on
  {C}omputational {L}inguistics}, pages 711$\,$--$\,$716, Kopenhagen, Denmark,
  1996.

\bibitem[Lewis and Crews(1985)]{lewis1985evolution}
Byron~C Lewis and Albert~E Crews.
\newblock The evolution of benchmarking as a computer performance evaluation
  technique.
\newblock \emph{MIS Quarterly}, pages 7--16, 1985.

\bibitem[Li(2019)]{FeiFeiLiWhere}
Fei-Fei Li.
\newblock {Fei-Fei Li}\,---\,{{Where Did ImageNet Come From}}?, 2019.
\newblock URL \url{https://www.youtube.com/watch?v=Z7naK1uq1F8}.

\bibitem[Liberman(2010)]{liberman2010fred}
Mark Liberman.
\newblock Fred {{Jelinek}}.
\newblock \emph{Computational Linguistics}, 36\penalty0 (4):\penalty0 595--599,
  2010.

\bibitem[Liu et~al.(2020)Liu, Ouyang, Wang, Fieguth, Chen, Liu, and
  Pietik{\"a}inen]{liu2018}
Li~Liu, Wanli Ouyang, Xiaogang Wang, Paul Fieguth, Jie Chen, Xinwang Liu, and
  Matti Pietik{\"a}inen.
\newblock Deep learning for generic object detection: A survey.
\newblock \emph{International Journal of Computer Vision}, 128\penalty0
  (2):\penalty0 261--318, 2020.

\bibitem[Malik(2020)]{malik2020hierarchy}
Momin~M. Malik.
\newblock A hierarchy of limitations in machine learning.
\newblock \emph{CoRR}, abs/2002.05193, 2020.
\newblock URL \url{https://arxiv.org/abs/2002.05193}.

\bibitem[McCann et~al.(2018)McCann, Keskar, Xiong, and
  Socher]{mccann2018natural}
Bryan McCann, Nitish~Shirish Keskar, Caiming Xiong, and Richard Socher.
\newblock The natural language decathlon: Multitask learning as question
  answering.
\newblock \emph{arXiv preprint arXiv:1806.08730}, 2018.

\bibitem[Merler et~al.(2019)Merler, Ratha, Feris, and
  Smith]{merler2019diversity}
Michele Merler, Nalini Ratha, Rogerio~S. Feris, and John~R. Smith.
\newblock Diversity in faces, 2019.
\newblock https://arxiv.org/abs/1901.10436.

\bibitem[Miller(1995)]{miller1995wordnet}
George~A Miller.
\newblock {WordNet:} {A} lexical database for {English}.
\newblock \emph{Communications of the ACM}, 38\penalty0 (11):\penalty0 39--41,
  1995.

\bibitem[Miller and Nicely(1955)]{Mil:Nic:55}
George~A Miller and Patricia~E Nicely.
\newblock An analysis of perceptual confusions among some {English} consonants.
\newblock \emph{The Journal of the Acoustical Society of America}, 27\penalty0
  (2):\penalty0 338--352, 1955.

\bibitem[Mitchell et~al.(2019)Mitchell, Wu, Zaldivar, Barnes, Vasserman,
  Hutchinson, Spitzer, Raji, and Gebru]{modelcards}
Margaret Mitchell, Simone Wu, Andrew Zaldivar, Parker Barnes, Lucy Vasserman,
  Ben Hutchinson, Elena Spitzer, Inioluwa~Deborah Raji, and Timnit Gebru.
\newblock Model cards for model reporting.
\newblock In \emph{Proceedings of the Conference on Fairness, Accountability,
  and Transparency}, FAT* '19, page 220–229, New York, NY, USA, 2019.
  Association for Computing Machinery.
\newblock ISBN 9781450361255.
\newblock \doi{10.1145/3287560.3287596}.
\newblock URL \url{https://doi.org/10.1145/3287560.3287596}.

\bibitem[Mitchell(2021)]{mitchell2021ai}
Melanie Mitchell.
\newblock Why {AI} is harder than we think.
\newblock \emph{arXiv preprint arXiv:2104.12871}, 2021.

\bibitem[Newell(1975)]{Newell:75}
Allen Newell.
\newblock A tutorial on speech understanding systems.
\newblock In D.~Raj Reddy, editor, \emph{Speech Recognition: {Invited} Papers
  Presented at the 1974 {IEEE} Symposium}, pages 4--54. Academic Press, New
  York, 1975.

\bibitem[Newell et~al.(1959)Newell, Shaw, and Simon]{newell1959report}
Allen Newell, John~C. Shaw, and Herbert~A. Simon.
\newblock Report on a general problem-solving program.
\newblock In \emph{Proceedings of the International Conference on Information
  Processing}, pages 256--264, 1959.

\bibitem[Niven and Kao(2019)]{Niv:Kao:19}
Timothy Niven and Hung-Yu Kao.
\newblock Probing neural network comprehension of natural language arguments.
\newblock In \emph{Proceedings of the 57th Annual Meeting of the Association
  for Computational Linguistics}, pages 4658--4664, Florence, Italy, 2019.
  Association for Computational Linguistics.
\newblock URL \url{https://www.aclweb.org/anthology/P19-1459}.

\bibitem[O'Leary-Kelly and Vokurka(1998)]{o1998empirical}
Scott~W O'Leary-Kelly and Robert~J Vokurka.
\newblock The empirical assessment of construct validity.
\newblock \emph{Journal of operations management}, 16\penalty0 (4):\penalty0
  387--405, 1998.

\bibitem[Paullada et~al.(2020)Paullada, Raji, Bender, Denton, and
  Hanna]{paullada2020data}
Amandalynne Paullada, Inioluwa~Deborah Raji, Emily~M Bender, Emily Denton, and
  Alex Hanna.
\newblock Data and its (dis)contents: {A} survey of dataset development and use
  in machine learning research.
\newblock \emph{arXiv preprint arXiv:2012.05345}, 2020.

\bibitem[Pennachin and Goertzel(2007)]{Pennachin2007}
Cassio Pennachin and Ben Goertzel.
\newblock Contemporary approaches to artificial general intelligence.
\newblock In Ben Goertzel and Cassio Pennachin, editors, \emph{Artificial
  General Intelligence}, pages 1--30. Springer Berlin Heidelberg, Berlin,
  Heidelberg, 2007.
\newblock ISBN 978-3-540-68677-4.
\newblock \doi{10.1007/978-3-540-68677-4_1}.
\newblock URL \url{https://doi.org/10.1007/978-3-540-68677-4_1}.

\bibitem[Phillips et~al.(2000)Phillips, Moon, Rizvi, and
  Rauss]{phillips2000feret}
P~Jonathon Phillips, Hyeonjoon Moon, Syed~A Rizvi, and Patrick~J Rauss.
\newblock The {FERET} evaluation methodology for face-recognition algorithms.
\newblock \emph{IEEE Transactions on pattern analysis and machine
  intelligence}, 22\penalty0 (10):\penalty0 1090--1104, 2000.

\bibitem[Przybocki et~al.(2009)Przybocki, Peterson, Bronsart, and
  Sanders]{przybocki2009nist}
Mark Przybocki, Kay Peterson, S{\'e}bastien Bronsart, and Gregory Sanders.
\newblock The {NIST} 2008 metrics for machine translation challenge—overview,
  methodology, metrics, and results.
\newblock \emph{Machine Translation}, 23\penalty0 (2):\penalty0 71--103, 2009.

\bibitem[Qiu et~al.(2020)Qiu, Sun, Xu, Shao, Dai, and Huang]{Qiu-etal2020}
XiPeng Qiu, TianXiang Sun, YiGe Xu, YiGe Shao, Ning Dai, and XuanJing Huang.
\newblock Pre-trained models for natural language processing: {A} survey.
\newblock \emph{{SCIENCE} {CHINA} Technological Sciences}, 63\penalty0
  (10):\penalty0 1872--1897, 2020.
\newblock URL \url{https://engine.scichina.com/publisher/Science China
  Press/journal/SCIENCE CHINA Technological
  Sciences/63/10/10.1007/s11431-020-1647-3}.

\bibitem[Radin(2017)]{radinDigitalNativesHow2017}
Joanna Radin.
\newblock ``{{Digital Natives}}'': {{How Medical}} and {{Indigenous Histories
  Matter}} for {{Big Data}}.
\newblock \emph{Osiris}, 32\penalty0 (1):\penalty0 43--64, September 2017.
\newblock ISSN 0369-7827, 1933-8287.
\newblock \doi{10.1086/693853}.

\bibitem[Raji and Buolamwini(2019)]{Raji2019}
Inioluwa~Deborah Raji and Joy Buolamwini.
\newblock Actionable auditing: Investigating the impact of publicly naming
  biased performance results of commercial {AI} products.
\newblock In \emph{Proceedings of the 2019 {AAAI/ACM} Conference on {AI},
  Ethics, and Society}, AIES '19, page 429–435, New York, NY, USA, 2019.
  Association for Computing Machinery.
\newblock ISBN 9781450363242.
\newblock \doi{10.1145/3306618.3314244}.
\newblock URL \url{https://doi.org/10.1145/3306618.3314244}.

\bibitem[Reddy(1979)]{Reddy:79}
Michael~J Reddy.
\newblock The conduit metaphor: A case of frame conflict in our language about
  language.
\newblock In Andrew Ortony, editor, \emph{Metaphor and Thought}, pages
  164--201. Cambridge University Press, 1979.

\bibitem[Ribeiro et~al.(2020)Ribeiro, Wu, Guestrin, and
  Singh]{ribeiro-etal-2020-beyond}
Marco~Tulio Ribeiro, Tongshuang Wu, Carlos Guestrin, and Sameer Singh.
\newblock Beyond accuracy: Behavioral testing of {NLP} models with
  {C}heck{L}ist.
\newblock In \emph{Proceedings of the 58th Annual Meeting of the Association
  for Computational Linguistics}, pages 4902--4912, Online, July 2020.
  Association for Computational Linguistics.
\newblock \doi{10.18653/v1/2020.acl-main.442}.
\newblock URL \url{https://www.aclweb.org/anthology/2020.acl-main.442}.

\bibitem[Russakovsky et~al.(2015)Russakovsky, Deng, Su, Krause, Satheesh, Ma,
  Huang, Karpathy, Khosla, Bernstein, Berg, and
  {Fei-Fei}]{russakovskyImageNetLargeScale2015}
Olga Russakovsky, Jia Deng, Hao Su, Jonathan Krause, Sanjeev Satheesh, Sean Ma,
  Zhiheng Huang, Andrej Karpathy, Aditya Khosla, Michael Bernstein,
  Alexander~C. Berg, and Li~{Fei-Fei}.
\newblock {{ImageNet Large Scale Visual Recognition Challenge}}.
\newblock \emph{International Journal of Computer Vision}, 115\penalty0
  (3):\penalty0 211--252, December 2015.
\newblock ISSN 0920-5691, 1573-1405.
\newblock \doi{10.1007/s11263-015-0816-y}.

\bibitem[Sambasivan et~al.(2021)Sambasivan, Kapania, Highfill, Akrong,
  Paritosh, and Aroyo]{sambasivan2021everyone}
Nithya Sambasivan, Shivani Kapania, Hannah Highfill, Diana Akrong, Praveen
  Paritosh, and Lora~M Aroyo.
\newblock “everyone wants to do the model work, not the data work”: Data
  cascades in high-stakes {AI}.
\newblock In \emph{Proceedings of the 2021 CHI Conference on Human Factors in
  Computing Systems}, pages 1--15, 2021.

\bibitem[Scheuerman et~al.(2021)Scheuerman, Hanna, and
  Denton]{scheuerman2021datasets}
Morgan~Klaus Scheuerman, Alex Hanna, and Emily Denton.
\newblock Do datasets have politics? {Disciplinary} values in computer vision
  dataset development.
\newblock \emph{Proceedings of the ACM on Human-Computer Interaction},
  5\penalty0 (CSCW2):\penalty0 1--37, 2021.

\bibitem[Schlangen(2021)]{Schlangen2020TargetingTB}
David Schlangen.
\newblock Targeting the benchmark: On methodology in current natural language
  processing research.
\newblock In \emph{Proceedings of the 59th Annual Meeting of the Association
  for Computational Linguistics and the 11th International Joint Conference on
  Natural Language Processing (Volume 2: Short Papers)}, pages 670--674,
  Online, August 2021. Association for Computational Linguistics.
\newblock \doi{10.18653/v1/2021.acl-short.85}.
\newblock URL \url{https://aclanthology.org/2021.acl-short.85}.

\bibitem[Schwartz et~al.(2020)Schwartz, Dodge, Smith, and
  Etzioni]{Schwartz2019}
Roy Schwartz, Jesse Dodge, Noah~A. Smith, and Oren Etzioni.
\newblock Green {AI}.
\newblock \emph{Commun. ACM}, 63\penalty0 (12):\penalty0 54–63, November
  2020.
\newblock ISSN 0001-0782.
\newblock \doi{10.1145/3381831}.
\newblock URL \url{https://doi.org/10.1145/3381831}.

\bibitem[Sculley et~al.(2018)Sculley, Snoek, Wiltschko, and
  Rahimi]{Sculley2018WinnersCO}
D.~Sculley, Jasper Snoek, Alexander~B. Wiltschko, and A.~Rahimi.
\newblock Winner's curse? {On} pace, progress, and empirical rigor.
\newblock In \emph{ICLR}, 2018.

\bibitem[Shankar et~al.(2017)Shankar, Halpern, Breck, Atwood, Wilson, and
  Sculley]{shankar2017no}
Shreya Shankar, Yoni Halpern, Eric Breck, James Atwood, Jimbo Wilson, and
  D~Sculley.
\newblock No classification without representation: Assessing geodiversity
  issues in open data sets for the developing world.
\newblock \emph{arXiv preprint arXiv:1711.08536}, 2017.

\bibitem[Shevlin et~al.(2019)Shevlin, Vold, Crosby, and Halina]{Shevlin2019}
Henry Shevlin, Karina Vold, Matthew Crosby, and Marta Halina.
\newblock The limits of machine intelligence.
\newblock \emph{Science \& Society}, 20\penalty0 (10), 2019.

\bibitem[Shotton et~al.(2006)Shotton, Winn, Rother, and
  Criminisi]{shotton2006textonboost}
Jamie Shotton, John Winn, Carsten Rother, and Antonio Criminisi.
\newblock Textonboost: Joint appearance, shape and context modeling for
  multi-class object recognition and segmentation.
\newblock In \emph{European conference on computer vision}, pages 1--15.
  Springer, 2006.

\bibitem[Simonite(2015)]{simonite2015and}
Tom Simonite.
\newblock Why and how {Baidu} cheated an artificial intelligence test.
\newblock \emph{MIT Technology Review}, 2015.

\bibitem[Stiles and Wilcox(1974)]{Grover:74}
Norman Stiles and Daniel Wilcox.
\newblock \emph{Grover and the Everything in the Whole Wide World Museum}.
\newblock Random House, New York, 1974.
\newblock Illustrations by Joe Mathieu.

\bibitem[Stitzlein(2004)]{stitzlein2004replacing}
Sarah~M Stitzlein.
\newblock Replacing the `view from nowhere': {A} pragmatist-feminist science
  classroom.
\newblock \emph{Electronic Journal of Science Education}, 9\penalty0 (2), 2004.

\bibitem[Strubell et~al.(2019)Strubell, Ganesh, and McCallum]{Strubell:2019}
Emma Strubell, Ananya Ganesh, and Andrew McCallum.
\newblock Energy and policy considerations for deep learning in {NLP}.
\newblock In \emph{Proceedings of the 57th Annual Meeting of the Association
  for Computational Linguistics}, pages 3645--3650, Florence, Italy, July 2019.
  Association for Computational Linguistics.
\newblock \doi{10.18653/v1/P19-1355}.
\newblock URL \url{https://www.aclweb.org/anthology/P19-1355}.

\bibitem[Sutskever(2018)]{Sutskever2018}
Ilya Sutskever.
\newblock Recent advances in deep learning and {AI} from {OpenAI}, 2018.
\newblock Keynote talk at {AI} Frontiers Conference.

\bibitem[Thomas and Uminsky(2020)]{thomas2020problem}
Rachel Thomas and David Uminsky.
\newblock The problem with metrics is a fundamental problem for {AI}, 2020.
\newblock https://arxiv.org/abs/2002.08512.

\bibitem[Torralba and Efros(2011)]{torralba2011unbiased}
Antonio Torralba and Alexei~A Efros.
\newblock Unbiased look at dataset bias.
\newblock In \emph{{CVPR} 2011}, pages 1521--1528. IEEE, 2011.

\bibitem[van Miltenburg et~al.(2021)van Miltenburg, Clinciu, Du{\v{s}}ek,
  Gkatzia, Inglis, Lepp{\"a}nen, Mahamood, Manning, Schoch, Thomson, and
  Wen]{van:Cli:Gka:21}
Emiel van Miltenburg, Miruna Clinciu, Ond{\v{r}}ej Du{\v{s}}ek, Dimitra
  Gkatzia, Stephanie Inglis, Leo Lepp{\"a}nen, Saad Mahamood, Emma Manning,
  Stephanie Schoch, Craig Thomson, and Luou Wen.
\newblock Underreporting of errors in {NLG} output, and what to do about it.
\newblock In \emph{Proceedings of the 14th International Conference on Natural
  Language Generation}, pages 140--153, Aberdeen, Scotland, UK, August 2021.
  Association for Computational Linguistics.
\newblock URL \url{https://aclanthology.org/2021.inlg-1.14}.

\bibitem[Voss(2007)]{Voss2007}
Peter Voss.
\newblock Essentials of general intelligence: The direct path to artificial
  general intelligence.
\newblock In Ben Goertzel and Cassio Pennachin, editors, \emph{Artificial
  General Intelligence}, pages 131--157. Springer Berlin Heidelberg, Berlin,
  Heidelberg, 2007.

\bibitem[Wagstaff(2012)]{Wagstaff:2012}
Kiri~L. Wagstaff.
\newblock Machine learning that matters.
\newblock In \emph{Proceedings of the 29th International Coference on
  International Conference on Machine Learning}, ICML'12, page 1851–1856,
  Madison, WI, USA, 2012. Omnipress.
\newblock ISBN 9781450312851.

\bibitem[Wang et~al.(2019a)Wang, Singh, Michael, Hill, Levy, and
  Bowman]{wang2018glue}
Alex Wang, Amanpreet Singh, Julian Michael, Felix Hill, Omer Levy, and Samuel
  Bowman.
\newblock {GLUE}: {A} multi-task benchmark and analysis platform for natural
  language understanding.
\newblock In \emph{7th International Conference on Learning Representations,
  ICLR 2019}, 2019a.

\bibitem[Wang et~al.(2019b)Wang, Pruksachatkun, Nangia, Singh, Michael, Hill,
  Levy, and Bowman]{wang2019superglue}
Alex Wang, Yada Pruksachatkun, Nikita Nangia, Amanpreet Singh, Julian Michael,
  Felix Hill, Omer Levy, and Samuel Bowman.
\newblock {SuperGLUE}: A stickier benchmark for general-purpose language
  understanding systems.
\newblock In \emph{Advances in Neural Information Processing Systems}, pages
  3266--3280, 2019b.

\bibitem[Waseem et~al.(2020)Waseem, Lulz, Bingel, and
  Augenstein]{waseem2020disembodied}
Zeerak Waseem, Smarika Lulz, Joachim Bingel, and Isabelle Augenstein.
\newblock Disembodied machine learning: {On} the illusion of objectivity in
  {NLP}, 2020.
\newblock https://openreview.net/forum?id=fkAxTMzy3fs.

\bibitem[Wetzel and Bond(2012)]{wetzel-bond-2012-enriching}
Dominikus Wetzel and Francis Bond.
\newblock Enriching parallel corpora for statistical machine translation with
  semantic negation rephrasing.
\newblock In \emph{Proceedings of the Sixth Workshop on Syntax, Semantics and
  Structure in Statistical Translation}, pages 20--29, Jeju, Republic of Korea,
  July 2012. Association for Computational Linguistics.
\newblock URL \url{https://www.aclweb.org/anthology/W12-4203}.

\bibitem[Yerushalmy(1947)]{Yerushalmy:47}
Jacob Yerushalmy.
\newblock Statistical problems in assessing methods of medical diagnosis, with
  special reference to {X-ray} techniques.
\newblock \emph{Public Health Reports (1896-1970)}, pages 1432--1449, 1947.

\bibitem[Zellers et~al.(2021)Zellers, Holtzman, Clark, Qin, Farhadi, and
  Choi]{zellers2020evaluating}
Rowan Zellers, Ari Holtzman, Elizabeth Clark, Lianhui Qin, Ali Farhadi, and
  Yejin Choi.
\newblock {T}uring{A}dvice: A generative and dynamic evaluation of language
  use.
\newblock In \emph{Proceedings of the 2021 Conference of the North American
  Chapter of the Association for Computational Linguistics: Human Language
  Technologies}, pages 4856--4880, Online, June 2021. Association for
  Computational Linguistics.
\newblock \doi{10.18653/v1/2021.naacl-main.386}.
\newblock URL \url{https://aclanthology.org/2021.naacl-main.386}.

\end{thebibliography}

\appendix
\section{Appendix A: Details of Alternative Evaluation Methods}
\label{appa}

\begin{wrapfigure}{l}{0.3\textwidth}
\scalebox{.20}{\includegraphics{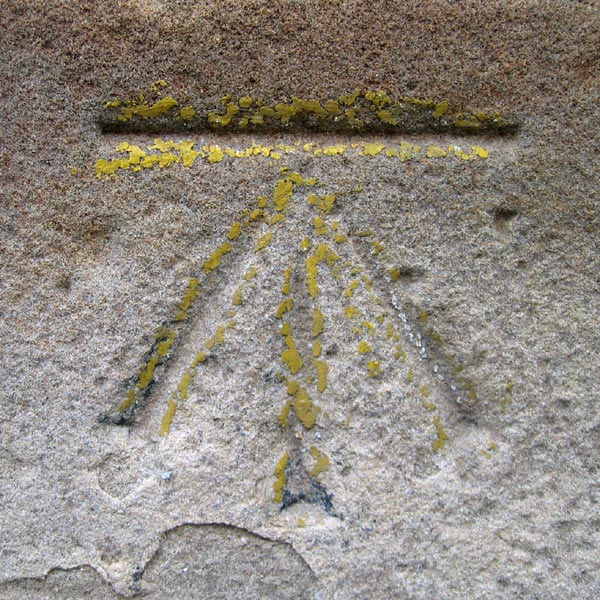}}
\caption{Photo of a benchmark in Edinburgh, by \href{https://commons.wikimedia.org/wiki/User:JeremyA}{Jeremy Atherton}, used without modification according to license  \href{https://creativecommons.org/licenses/by-sa/2.5/}{CC-BY-SA-2.5}}
\label{fig:benchmark}
\end{wrapfigure}

The etymology for the term ``benchmark'' refers to a mark that was added to buildings to indicate the position of a surveyor's bench (see Fig.~\ref{fig:benchmark}), itself a tool for creating a level surface on which to put a leveling rod, used in the process of surveying. This etymological source contrasts with the use of benchmarks for charting the state of the art: surveyors are not in the business of measuring the furthest anyone has gone along some particular trail but rather in understanding the shape of the landscape and how it has changed over time \citep{kahmen1988surveying}. 

We can take further inspiration from the field of surveying, as we think about how the measurements we take relate to the terrain we wish to understand: ``A surveyor is not only charged with providing results derived from [their] measurements, but also has to give an indication of the quality and reliability of these. This requires a clear understanding of the functional and stochastic relationships between measured quantities and derived results, as well as a solid understanding of the external factors that influence the measurements.'' \citep[p.1]{kahmen1988surveying}. 

We describe in this appendix other methodologies that can move us towards the perhaps more worthy goal of filling in our picture of the landscape. In turn, we provide an overview of alternative, underexplored evaluation techniques including testsuites, audits and adversarial testing; system output analysis; ablation testing; and analysis of model properties.  

\subsection{Testsuites, Audits and Adversarial Testing}
\label{sec:ts}

Typical benchmark evaluation datasets are sampled from some larger dataset (e.g.\ via a train/test split) such that the frequency distribution of test item \emph{types} in the test data is influenced by their distribution in that underlying dataset. In contrast, testsuites and audits specifically design their test sets to map out some space of test item types and evaluate systems in terms of the extent to which they can handle them.\footnote{A notable exception to the trend of benchmarks to not include test suites is GLUE, which includes among its component tests a testsuite mapping out various linguistic constructions in English \citep{wang2018glue}.}  The testsuite-based approach has a long history in NLP, with notable early publications including \citet{TSNLP} and recent work such as \citet{ribeiro-etal-2020-beyond}.

There are various approaches to the design of such test suites. A more audit-like methodology, exemplified by \citet{GenderShades}, creates test sets balanced for sensitive categories so as to be able to test for differential performance across those categories. On the other hand, explicit adversarial testing seeks to explore the edges of a system's competence by finding minimally contrasting pairs of examples where the system being tested succeeds on one member of the pair and fails on the other \citep{ettinger-etal-2017-towards}.\footnote{Adversarial testing also includes work like \citet{Niv:Kao:19} that discovers what kind of spurious cues systems are leveraging to effectively ``cheat'' on a particular benchmark and creates alternate versions of the testsets that neutralize those cues.} Importantly, these evaluation approaches are designed around diagnosing particular areas of system failure: the point of the tests isn't to show which systems can ``solve'' them but to understand which aspects of the problem space remain challenging and represent the operational limits in deployment.

\subsection{System Output Analysis}
\label{sec:output}

System output analysis is another way to explore the system output in detail. This can take the form of error analysis, disaggregated analysis, and counterfactual analysis.

\paragraph{Error analysis} Error analysis involves the detailed analysis of system errors, by either mechanically or manually inspecting system inputs. Mechanical analysis can include simple techniques such as developing a confusion matrix in labeling tasks. A confusion matrix compares expected labels to system output labels and provides a summary of which categories are most reliably labeled and which are most frequently confused for each other. (This is an old technique, initially used in the study of human phonetic perception \citep{Mil:Nic:55} and related to even earlier work comparing medical diagnostics when lacking access to ground truth \citep{Yerushalmy:47}.) Other mechanical analyses include looking at errors by easily automatically measurable properties of system input such as sentence length or presence or number of out of vocabulary items. 

More detailed error analysis digs into the specific system inputs to look for patterns that can't necessarily be measured automatically and might not be modeled in any way by the system being evaluated. For example, error analysis of a sentiment analysis system might find that it is frequently tripped up by sarcasm or in a simpler case by sentences with phenomena such as coordination or subordinate clauses. Error analysis of a machine translation (MT) system might find that it frequently fails on examples with negation \citep{wetzel-bond-2012-enriching,fancellu-webber-2015-translating-negation,hossain2020its}. Error analysis can turn up important problems that don't have a large effect on the metric. For example, metrics for MT systems (including but not limited to BLEU) are also not good at measuring the impact of negation errors \citep{hossain2020its}.

\paragraph{Disaggregated analysis}
\label{sec:disagg}

Disaggregated analysis can reveal disparate patterns of performance that may not be visible through aggregate metrics alone. This method has been leveraged within the ground-breaking audit of facial analysis systems, performed by \cite{GenderShades}, that evaluated performance across unitary and intersectional subgroups defined by gender and Fitzpatrick skin type. This analysis revealed significant disparities in model performance\,---\,with darker-skinned female subjects experiencing the highest error rates\,---\,that was not visible through examination of aggregate performance metrics alone. The method of disaggregated analysis has since been adopted by a myriad of auditing and evaluation works \citep[e.g.][]{Raji2019}  and integrated into frameworks of standardized model reporting \citep{modelcards}. Following these works, we encourage researchers to report performance metrics on socially salient slices of their dataset, in addition to the full test set. 

\paragraph{Counterfactual analysis}

Counterfactual analysis is another technique of model evaluation and assessment that has gained in popularity in recent years. At a high level, these methods evaluate how a model's output changes in response to a counterfactual change in the input. This method has been leveraged for fairness-informed analysis of natural language processing systems by comparing model performance on paired inputs that differ only in a reference to a sensitive identity group \citep{Garg2019, Hutchinson2020}. While both counterfactual analysis and disaggregated analysis have been leveraged to disparities in model performance for different sensitive groups, counterfactual analysis can additionally provide insight into causal mechanisms underlying particular patterns in performance. Counterfactual analysis can also be leveraged to assess model robustness to small distribution shifts \citep{Christensen2019}.

\smallskip

The results of system output analysis tend to be rich and detailed and not amenable for quick cross-system comparison. But this is a feature, in our view, and not a bug: the goal, after all, is not anoint one system the winner (until some new system claims that spot), but rather to understand how aspects of system design map onto different aspects of the problem space so as to inform the next iteration of system development.

\subsection{Ablation Testing}
\label{sec:ablation}

Another well-established technique for understanding system performance is ablation testing. Ablation testing, as named by \citet{Newell:75} by analogy with similar studies in biology, involves isolating the contributions of different system components by removing them, one by one, and evaluating the modified system. In statistical NLP prior to deep learning approaches, ablation testing was commonly applied to different feature sets to explore the extent to which systems were using information captured by different aspects of input representation. Ablation testing can also be performed on subsets of training data to explore the effect of e.g.\ in-domain v.\ out-of-domain training data or on components of system architecture, to the extent that these can be removed without completely disabling the system. \citet{Heinzerling:19}, in a discussion inspired by \cite{Niv:Kao:19}, suggests various data ablations that can be applied to test data as well, to investigate what cues a system might be relying on.

\subsection{Analysis of Model Properties}
\label{sec:mp}

System performance on test items (in aggregate as in test sets in standard benchmarks or in detail as in error analysis or test suites or audits) is only one facet of systems to consider, especially when trying to gauge which approaches are most feasible for practical applications. Other dimensions include energy consumption (both for development and testing) \citep{Strubell:2019,Henderson:2020, Schwartz2019,Ethayarajh2020}; memory and compute requirements, which may be more or less constrained depending on the deployment context \citep{Ethayarajh2020, dodge2019}; and stability in the face of perturbations to the training data \citep{Sculley2018WinnersCO}. This latter is especially important for systems that need to be continually retrained in order to handle changing input data, such as named entity recognition system that need to keep up with different public figures of note.

\end{document}